\documentclass[letterpaper, 10 pt, conference]{ieeeconf}  

\IEEEoverridecommandlockouts                              

\usepackage[pdftex]{graphicx}
\usepackage{amsmath}
\usepackage{amssymb}
\usepackage{diagbox}
\usepackage{makecell}
\usepackage{mathtools} 
\usepackage{url}
\usepackage{caption}
\usepackage{cite}
\usepackage{bm}
\makeatletter
\let\NAT@parse\undefined
\makeatother
\usepackage[colorlinks,bookmarks=false,citecolor=black,linkcolor=black,urlcolor=softblue]{hyperref}
\usepackage{algorithm}
\usepackage{algpseudocode}

\newcommand{\norm}[1]{\Vert{#1}\Vert}

\graphicspath{{./figures/}}
\DeclareGraphicsExtensions{.png,.jpg,.eps,.pdf}
\title{\LARGE \bf
  A Precise Real-Time Force-Aware Grasping System for Robust Aerial Manipulation
}

\author{Kenghou Hoi$^{1,2,*}$, Yuze Wu$^{1,2,*}$, Annan Ding$^{3,4,*}$, Junjie Wang$^{1,2}$, Anke Zhao$^{1,2}$,\\ Jialiang Hou$^{1,2,\dagger}$, Chengqian Zhang$^{3,4,\dagger}$, and Fei Gao$^{1,2,\dagger}$
\thanks{\textsuperscript{*}Equal contribution}
\thanks{$^{\dagger}$Corresponding author}
\thanks{\textsuperscript{1}State Key Laboratory of Industrial Control Technology, Zhejiang University, Hangzhou 310027, China.}
\thanks{\textsuperscript{2}Huzhou Institute, Zhejiang University, Huzhou 313000, China.}	
\thanks{\textsuperscript{3}State Key Laboratory of Fluid Power and Mechatronic Systems, School of Mechanical Engineering, Zhejiang University, Hangzhou 310058, China.}
\thanks{\textsuperscript{4}Zhejiang Key Laboratory of Additive Manufacturing Technology and Equipment, School of Mechanical Engineering, Zhejiang University, Hangzhou 310058, China.}
\thanks{E-mail: kenghouhoi@zju.edu.cn, jlhou8@gmail.com, fgaoaa@zju.edu.cn}
}

\begin{document}

	\makeatletter
	\let\@oldmaketitle\@maketitle
	\renewcommand{\@maketitle}{\@oldmaketitle
		\begin{center}
			\includegraphics[width=1.0\linewidth]{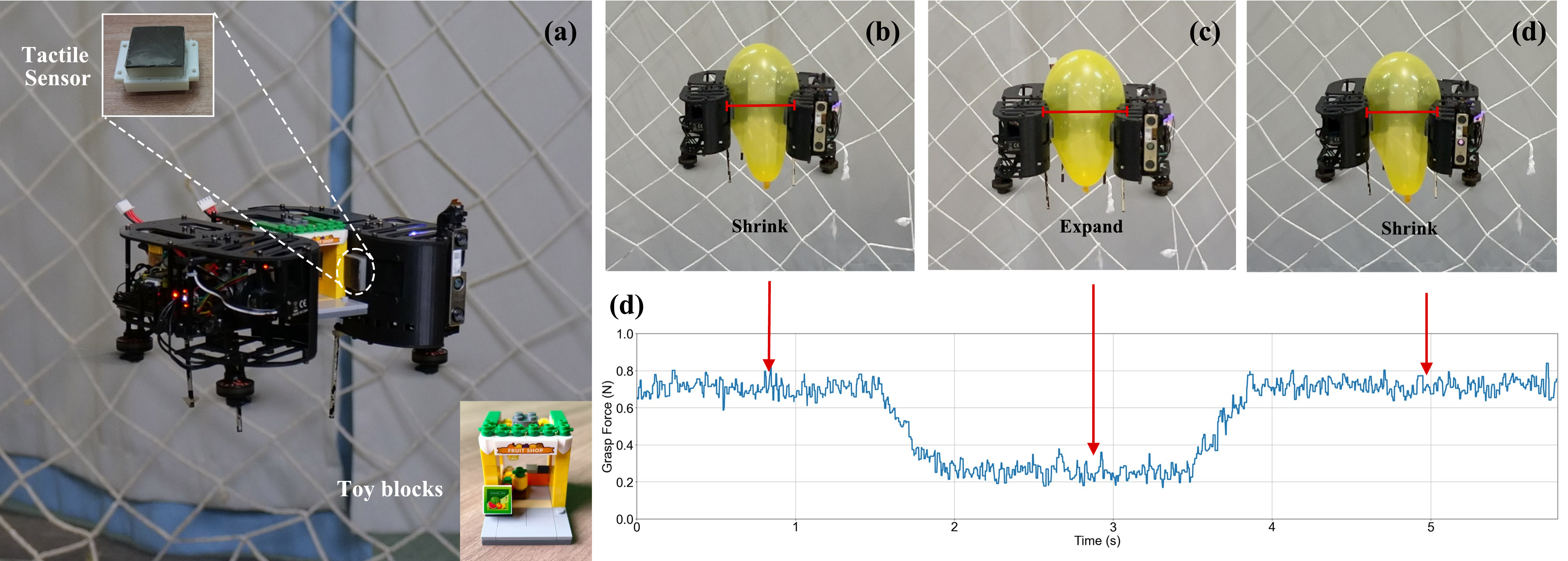}
		\end{center}
		\captionsetup{font={small}}
        \captionof{figure}{
        \label{fig:top}
        Grasping experiments. (a) Experimental setup for handling fragile toy blocks using our proposed magnetic soft tactile sensor, demonstrating gentle manipulation capability. (b-d) Deformable quadrotor sequence showing shrinkage-expansion-shrinkage states, with (c) exhibiting wider balloon deformation during expansion compared to shrunk configurations. (d) Measured grasping force from proposed tactile sensors, demonstrating force reduction during expansion and subsequent increase upon re-shrinkage.
        \vspace{-0.4cm}
        }
	}
	\makeatother
	\maketitle
	\setcounter{figure}{1}
	\thispagestyle{empty}
	\pagestyle{empty}

\definecolor{softblue}{RGB}{70, 130, 180}
\begin{abstract}
Aerial manipulation requires force-aware capabilities to enable safe and effective grasping and physical interaction. Previous works often rely on heavy, expensive force sensors unsuitable for typical quadrotor platforms, or perform grasping without force feedback, risking damage to fragile objects. To address these limitations, we propose a novel force-aware grasping framework incorporating six low-cost, sensitive skin-like tactile sensors. We introduce a magnetic-based tactile sensing module that provides high-precision three-dimensional force measurements. We eliminate geomagnetic interference through a reference Hall sensor and simplify the calibration process compared to previous work. The proposed framework enables precise force-aware grasping control, allowing safe manipulation of fragile objects and real-time weight measurement of grasped items. The system is validated through comprehensive real-world experiments, including balloon grasping, dynamic load variation tests, and ablation studies, demonstrating its effectiveness in various aerial manipulation scenarios. Our approach achieves fully onboard operation without external motion capture systems, significantly enhancing the practicality of force-sensitive aerial manipulation. The supplementary video is available at: \href{https://www.youtube.com/watch?v=mbcZkrJEf1I}{https://www.youtube.com/watch?v=mbcZkrJEf1I}.
\end{abstract}

\begin{figure*}[t]
	\centering
    \includegraphics[width=\linewidth]{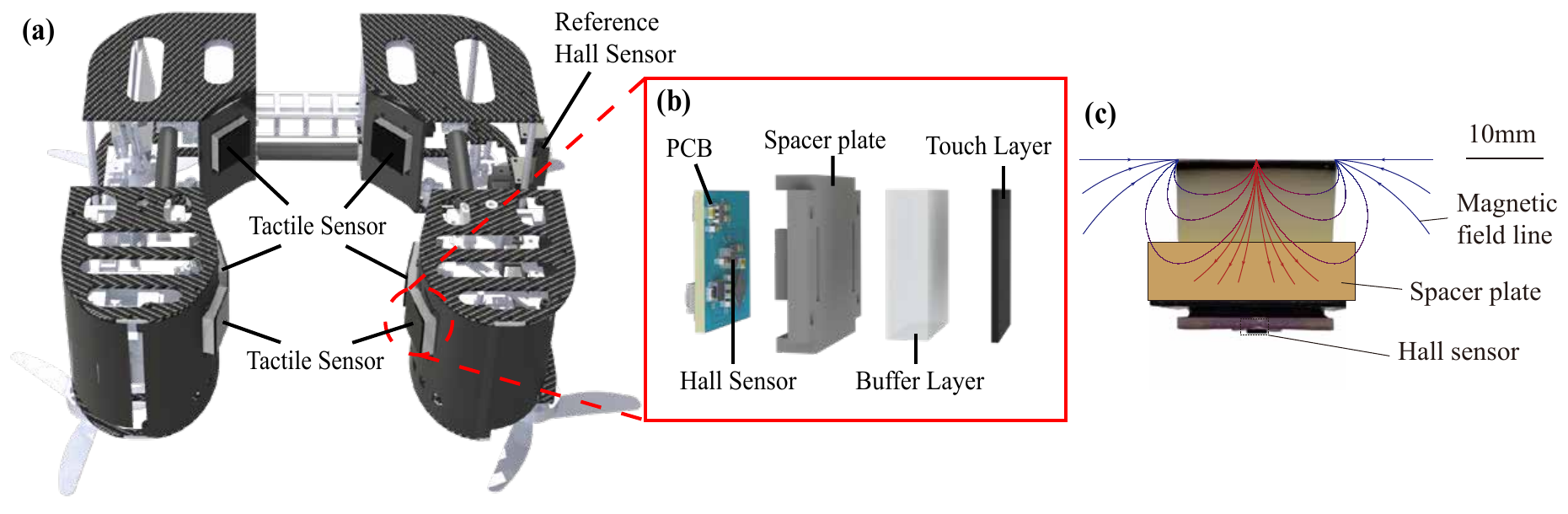}
	\caption{Hardware architecture. (a) Deformable quadrotor with six tactile sensors mounted on the inner wall.  (b) Exploded view of the proposed tactile sensor showing its internal components and assembly. (c) Working principle schematic of the magnetic soft tactile sensor.}
	\label{fig:robot_design}
    \vspace{-0.4cm}
\end{figure*}

\section{INTRODUCTION}
Aerial manipulation, by leveraging the inherent spatial maneuverability of aerial platforms, can realize physical interactions in complex workspaces inaccessible to ground robots. In recent years, a growing number of applications such as aerial grasping \cite{appius2022raptor, ubellacker2024high, xu2023aerial, zhao2023versatile, wu2023ring}, aerial writing \cite{guo2024flying, tzoumanikas2020aerial, bodie2019omnidirectional} and aerial interaction \cite{liang2023adaptive, benzi2022adaptive} have emerged, showing the promising potential for broadening the scope of manipulation area. However, bringing the aerial manipulation to the practice still faces significant challenges. On the one hand, the substantial mass and size of conventional manipulators severely degrade the agility and maneuverability of the whole aerial manipulation system, thereby hindering its deployment in confined indoor spaces or complex unstructured environments. On the other hand, most existing methods do not consider fragile and contact-rich tasks, such as grasping fragile objects. This limitation primarily stems from the absence of active perception of contact forces, a capability that is essential for estimating physical interaction and ensuring successful task execution. In this way, the development of a lightweight and compact aerial manipulation system capable of precise and force-aware grasping remains a significant open challenge.

At the heart of this challenge lies the accurate and real-time perception of contact forces. Previous works employ force and torque sensors at the end-effector to directly measure interaction forces \cite{guo2024flying, tzoumanikas2020aerial}. This approach is favored as the sensors offer high-precision measurements at the point of contact. However, such sensors entail a high cost and add considerable mass to the aerial platform. Another much cheaper and widely used one is the visual-tactile sensor that infers forces from the visual deformation of an elastic structure. While most methods \cite{lin20239dtact, bhirangi2024anyskin} obtain the pleasant measurements through data-driven methods, the sensor’s accuracy depends on visual quality. Thus, its performance can be compromised by visual noise and exhibits low sensitivity where small changes in contact force fail to produce a discernible visual signal. Besides, as most physical interactions involve complex contact patterns distributed over multiple points or surfaces, such methods suffer from computational burdens for simultaneously computing multiple forces with limited onboard computing resources. In contrast, the magnetic tactile sensors offer a potential solution to this dilemma. The sensor developed from previous work \cite{dai2024split} offers the highly precise three-dimensional force measurement under theoretical guarantee compared to others \cite{lin20239dtact, bhirangi2021reskin, bhirangi2024anyskin, lin2025pp}. The intrinsic magnetic sensing principle ensures robustness against optical disturbances and provides the high-resolution force feedback crucial for dynamic interaction. Furthermore, this sensor design is not only lightweight and compact, but also features a closed-form analytical model that guarantees real-time performance with minimal computational overhead, which makes it ideal for aerial manipulatio n tasks.

In this paper, we present a novel aerial manipulation system featuring a lightweight hand-like flying robot (weighing 556g) \cite{wu2026hand} integrated with a custom-designed magnetic soft tactile sensor based on our previous work \cite{dai2024split}, enabling fragile and precise manipulation during contact-rich aerial manipulation tasks. The lightweight magnetic soft tactile sensor units are arranged into a distributed array within the chassis of the quadrotor platform, providing rich spatial feedback with negligible impact on the quadrotor flight performance. When mounted on the dynamic aerial platform, the sensor's measurements are continuously corrupted by the Earth's geomagnetic field, as the changing attitude of the quadrotor translates this static field into large, unpredictable signal fluctuations. To decouple the force-induced signal from this environmental noise, we add a new Hall sensor on the platform to exclusively monitor the Earth's geomagnetic field, enabling the extraction of the true contact signal. By leveraging the filtered contact forces as an additional observation signal, the novel adaptive controller is designed to refine actuator commands, thereby enabling precise force modulation and safe interactions with fragile objects. In particular, the system demonstrates robust performance in handling objects with varying load conditions, ensuring accurate position control even under dynamic disturbances. Extensive physical experiments validate the effectiveness of our approach in achieving accurate and efficient force-aware grasping.

The main contributions of this work are summarized as follows:

\begin{itemize}
\item  A deformable quadrotor platform integrated with a lightweight, flexible, and precise tactile sensing array for diverse physical interactions.
\item An adaptive control framework combining admittance control and real-time force feedback for robust manipulation.
\item Extensive experimental validation demonstrating precise and efficient force-aware grasping across diverse objects and dynamic load conditions.
\end{itemize}

\section{RELATED WORKS}

\begin{figure*}[t]
\centering
    \includegraphics[width=\linewidth]{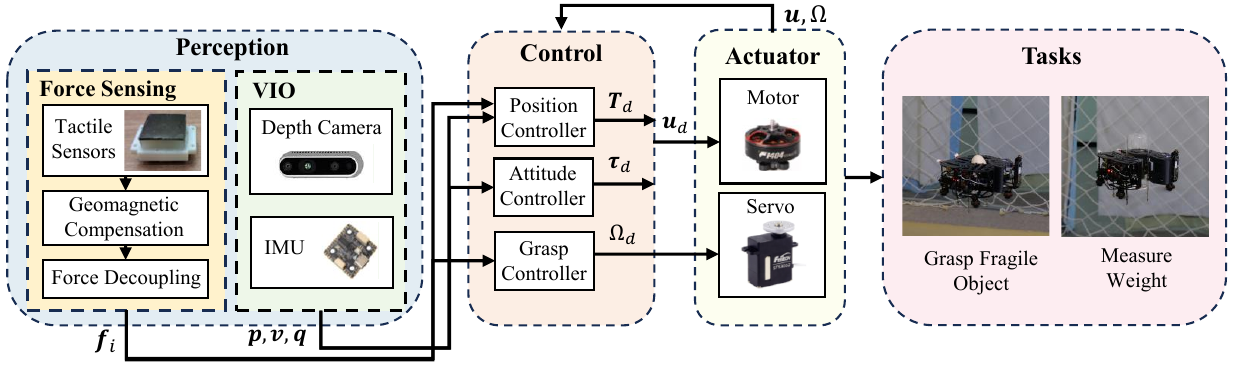}
    \caption{System overview showing the integration of perception (force sensing and visual-inertial odometry), control (position, attitude, and grasp controllers), actuation (motor and servo system), and task execution (fragile object grasping and weight measurement capabilities).}
    \vspace{-0.4cm}
\label{fig:pipeline}
\end{figure*}

\subsection{Manipulation with Tactile Sensor}
Tactile sensors are widely used for contact force perception in robotic manipulation tasks. Huang et al. \cite{huang20243d} employ a $16\times 16$ pressure sensor array to accomplish tasks such as Hex Key Collection, while Bhirangi et al. \cite{bhirangi2024anyskin} utilize a sensor based on \cite{bhirangi2021reskin} to achieve Plug Insertion. Lin et al. \cite{lin2025pp} implement Paper Picking with dexterous robotic hands using a visual-tactile sensor, and Xue et al. \cite{xue2025reactive} realize contact-rich tasks including wiping a vase and peeling a cucumber using GelSight \cite{yuan2017gelsight}.
However, these approaches have notable limitations: the sensor in \cite{huang20243d} only measures one-dimensional normal force, and vision-based tactile sensors like 9DTact \cite{lin20239dtact} often suffer from lower sampling rates and higher computational demands, making them less suitable for real-time fine manipulation. Magnetic-based tactile sensor ReSkin \cite{bhirangi2021reskin} and AnySkin \cite{bhirangi2024anyskin} did not thoroughly investigate the physical mechanism behind magnetic decoupling or the design of magnetic fields, primarily relying on data-driven calibration to achieve magnetic field-to-force mapping perception, thus lacking support from theoretical modeling. In contrast, our magnetic-based tactile sensing system provides high-bandwidth three-dimensional force measurements with greater sensitivity and accuracy than the Reskin series or visual-tactile sensors, without requiring extensive data collection or machine learning training.

\begingroup
\renewcommand{\arraystretch}{1.0}
\setlength{\tabcolsep}{4pt}
\begin{table}[t]
    \centering
    \caption{Comparison between sensor}
    \label{tab:sensor_comparison}
    \begin{tabular}{c|c|c|c|c}
        \hline
        \textbf{Characteristic} & F/T Sensor & 9DTact\cite{lin20239dtact} & ReSkin\cite{bhirangi2021reskin} & Ours\\
        \hline
        Frequency [Hz] & $1000$ & $90$ & $400$ & $50$\\
        Weight [g]& 280 & 20 & Not reported & 8\\
        Cost [\$] & 2000 & 15 & 30 & 10\\
        \makecell{Requires Training} & No & Yes & Yes & No\\
        Error [N] & $<0.01$ & $<0.4$ & $<0.3$ & $<0.05$ \\
        \hline
    \end{tabular}
\vspace{-1em}
\end{table}
\endgroup

\subsection{Aerial Manipulation}
Significant progress has been made in aerial manipulation across various applications. Guo et al. \cite{guo2024flying} used Force/Torque (F/T) sensors to achieve contact-aware motion and force planning for physical interaction. Lin et al. \cite{lin2025float} designed a fully actuated aerial robot capable of performing tasks such as watering plants and pulling curtains through attitude variation. Wu et al. \cite{wu2023ring} developed a retractable ring-shaped quadrotor that adjusts its frame size to grasp objects, while Ubellacker et al. \cite{ubellacker2024high} demonstrated high-speed aerial grasping using soft fingers with onboard perception. Wang et al. \cite{wang2025safe} employed multi-robot systems for agile transportation of cable-suspended payloads, and Li et al. \cite{li2025aerothrow} implemented an autonomous aerial throwing system using a Delta arm manipulator. However, most existing systems either employ rigid grippers that limit compliance during interaction or lack integrated high-resolution force sensing capabilities for fragile manipulation tasks. Our work distinguishes itself by integrating a deformable quadrotor platform with a distributed array of soft tactile sensors, enabling not only adaptive grasping but also real-time admittance force control specifically designed for fragile object manipulation. This combination of deformable structure, distributed tactile sensing, and integrated admittance control provides an effective approach toward practical aerial manipulation in real-world scenarios where precise force interaction is required.

\section{SYSTEM OVERVIEW}
Our quadrotor builds upon previous work \cite{wu2026hand} as a hand-like autonomous flying robot. It achieves aerial grasping and transportation through shape transformation controlled by a single actuator. As shown in Fig. \ref{fig:robot_design}, six tactile sensors are mounted on the inner wall of the central frame. The overall system architecture, illustrated in Fig. \ref{fig:pipeline}, comprises three main modules: perception, control and actuator. First, we developed a tactile sensing system capable of decoupling and estimating three-dimensional contact forces. Second, utilizing this force feedback in combination with positioning data, we design an adaptive controller to maintain stable hovering while achieving desired grasping forces. Finally, the quadrotor coordinates a servo motor and propulsion systems to execute precise force-aware aerial grasping operations. The system uses an onboard stereo camera for visual perception and an inertial measurement unit for state estimation, enabling fully autonomous operation without external motion capture systems. 

\section{TACTILE FORCE SENSING MODULE}
The proposed tactile sensing module operates based on magnetic field variations caused by deformation of a custom-designed flexible magnetic film. When external forces are applied, the displacement and tilt of this magnetic film modify the surrounding magnetic field distribution. These subtle magnetic changes are detected by a highly sensitive Hall-effect sensor integrated into the design. Through the developed decoupling algorithm, these magnetic variations are accurately converted into three-dimensional force measurements, providing high-resolution tactile perception capabilities. The compliant and soft physical structure of the sensor ensures safe interaction with objects while maintaining mechanical robustness, making it particularly suitable for fragile aerial manipulation applications where gentle contact is essential.

\begin{figure}[t]
	\centering
    \includegraphics[width=\linewidth]{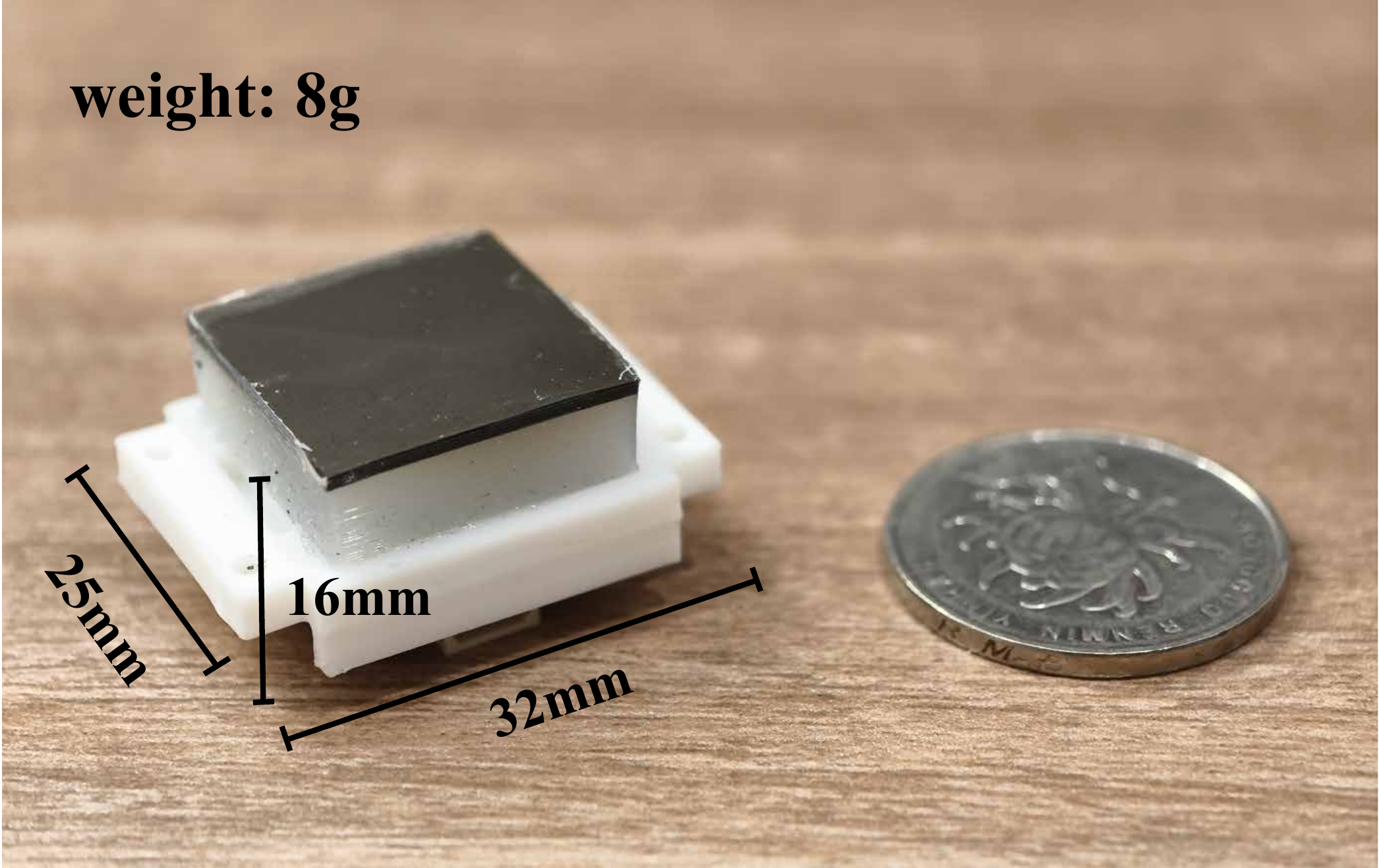}
    \caption{Sensor size comparison showing dimensions along all each axes with a coin for scale. The complete sensor weighs only 8 grams.}
	\label{fig:sensor_size}
    \vspace{-0.5cm}
\end{figure}

\subsection{Hardware Design}
Based on previous work \cite{dai2024split}, we designed a coin-sized sensor with overall dimensions of $25\times 32\times 16$ mm. The sensor consists of three parts: a silicone layer covered with a magnetic film, a 3D-printed component which used to fix the silicone and the sensor, and a custom Printed Circuit Board (PCB) with a Hall sensor. We mixed neodymium-iron-boron (NdFeB) particles with a flexible substrate material, followed by high-temperature curing and cutting into the desired shape. Referring to the origami technique, the composite is folded and magnetized to endow it with a specific centripetal magnetic field distribution. To ensure the minimum decoupling distance, we selected a silicone block with dimensions of $20\times 20\times 7$ mm, resulting in a 10 mm distance between the magnetic film and the sensor. During fabrication, Ecoflex 00-30 serves as the base material, while silicone oil (PMX-200; viscosity 5 cSt) or fumed silica nanoparticles are mixed into the silicone to tune its elastic modulus, thereby adjusting the sensor's measurement range and sensitivity. As the mass ratio of silicone oil to silicone increases, the elastic modulus of the soft silicone elastomer decreases correspondingly. 

To simplify wiring and facilitate communication with the host computer, we designed a compact circuit board (20 mm $\times$ 20 mm $\times$ 1.8 mm) integrating an \textit{STM32F042G6U6} as the main controller and an \textit{MLX90393SLW} Hall sensor chip. The board provides both CAN and UART interfaces. By leveraging CAN communication characteristics, we implemented daisy-chained data transmission and power delivery, significantly reducing wiring complexity in sensor array configurations. In this design, the UART interface of any individual sensor can access data from all sensors on the CAN bus, facilitating efficient data acquisition by the host computer. A custom communication protocol is developed to ensure optimal data transfer efficiency and real-time performance.

\subsection{Force Measuring}
Previous works \cite{dai2024split} have theoretically demonstrated the decoupling feasibility of sinusoidally magnetized films with an infinite period. While this approach provides important theoretical insights, practical manufacturing limitations prevent the realization of ideal magnetic field distributions. Based on this theory \cite{dai2024split}, we adopt a simplified centripetally magnetized magnetic field model to address these implementation constraints. The magnetic flux density measured by the Hall sensor is defined as $\bm{B}=[B_x,B_y,B_z]^\top$. To compensate for the errors caused by this simplified magnetic field, we introduce two compensation factors $B_{z}^{'}$ and $B_{z}^{''}$ to modify the theoretical model $B_{z}$ component. 
\begin{equation}
  B_{z}^{'}={k_1}B_z+c_1, B_{z}^{''}={k_2}B_z+c_2,
  \label{eq:z_compensate}
\end{equation} 
where $k_1, k_2, c_1, c_2 \in \mathbb{R}$ are compensation coefficients used to account for non-idealities in the magnetic field distribution. Since the shear modulus and elastic modulus are influenced by the combined effect of the magnetic film, silicone rubber elastomer, and adhesive, with potential defects introduced during manufacturing, practical implementation requires calibration based on silicone contact area, parameters $\bm{a}_i$ related to the shear modulus, elastic modulus, wavenumber of the ideally magnetized film, and elastomer thickness, along with position installation deviation compensation parameters $\bm{b}_i$. The three-dimensional force decoupling model for the $i$-th sensor is:
\begin{equation}
  \bm{f}_i = \bm{a}_{i} \odot \bm{S}(\bm{B}_i)+\bm{b}_{i},
  \quad \bm{f}_i, \bm{a}_i, \bm{b}_i, \bm{S}, \bm{B}_i \in \mathbb{R}^3,
  \label{eq:force}
\end{equation}
where
\begin{align}
\bm{S(B)} =
\begin{bmatrix}
  \arctan\!\left(\frac{B_x}{B_{z}^{'}-\tfrac{B_{z}^{'2}-(B_y^2-B_x^2)}{2B_{z}^{'}}}\right) \\[6pt]
  \arctan\!\left(\frac{B_y}{B_{z}^{'}-\tfrac{B_{z}^{'2}+(B_y^2-B_x^2)}{2B_{z}^{'}}}\right) \\[6pt]
  \ln\!\Biggl(\sqrt{\Bigl(B_{z}^{''}-\tfrac{B_{z}^{''}+(B_y^2-B_x^2)}{2B_{z}^{''}}\Bigr)^{2}+B_y^2}\,\Biggr)
\end{bmatrix}.
\label{eq:decoupling}
\end{align}

\subsection{Geomagnetic Compensation}
Compared with ground robots and manipulators, quadrotors exhibit higher maneuverability, but the proposed sensor is susceptible to geomagnetic interference, causing the magnetic flux density measurements to vary with attitude. To address this, we incorporated a reference Hall sensor fixed to the quadrotor body to measure the geomagnetic intensity in the body frame. This allows compensation of geomagnetic effects on the sensor measurements:
\begin{equation}
  \bm{\hat{B_i}}=\bm{B}_i-^{\mathcal{S}_i}_{\mathcal{S}_{ref}}\bm{R} \bm{B}_{ref},
  \label{eq:mag_compensate}
\end{equation}
where $\bm{B}_{ref} \in \mathbb{R}^3$ is the magnetic flux density measured by the reference Hall sensor, $^{\mathcal{S}_i}_{\mathcal{S}_{ref}}\bm{R} \in \mathbb{R}^{3\times 3}$ is the rotation matrix from the reference sensor to the $i$-th sensor, and finally we can obtain the corrected $\bm{\hat{B_i}} \in \mathbb{R}^3$. The corresponding compensated force estimate is:
\begin{align}
\hat{\bm{f}_i} = \bm{a}_{i} \odot \bm{S}(\hat{\bm{B}_i})+\bm{b}_{i}.
\label{eq:force_compensate}
\end{align}

\begin{figure}[h]
\centering
\includegraphics[width=\linewidth]{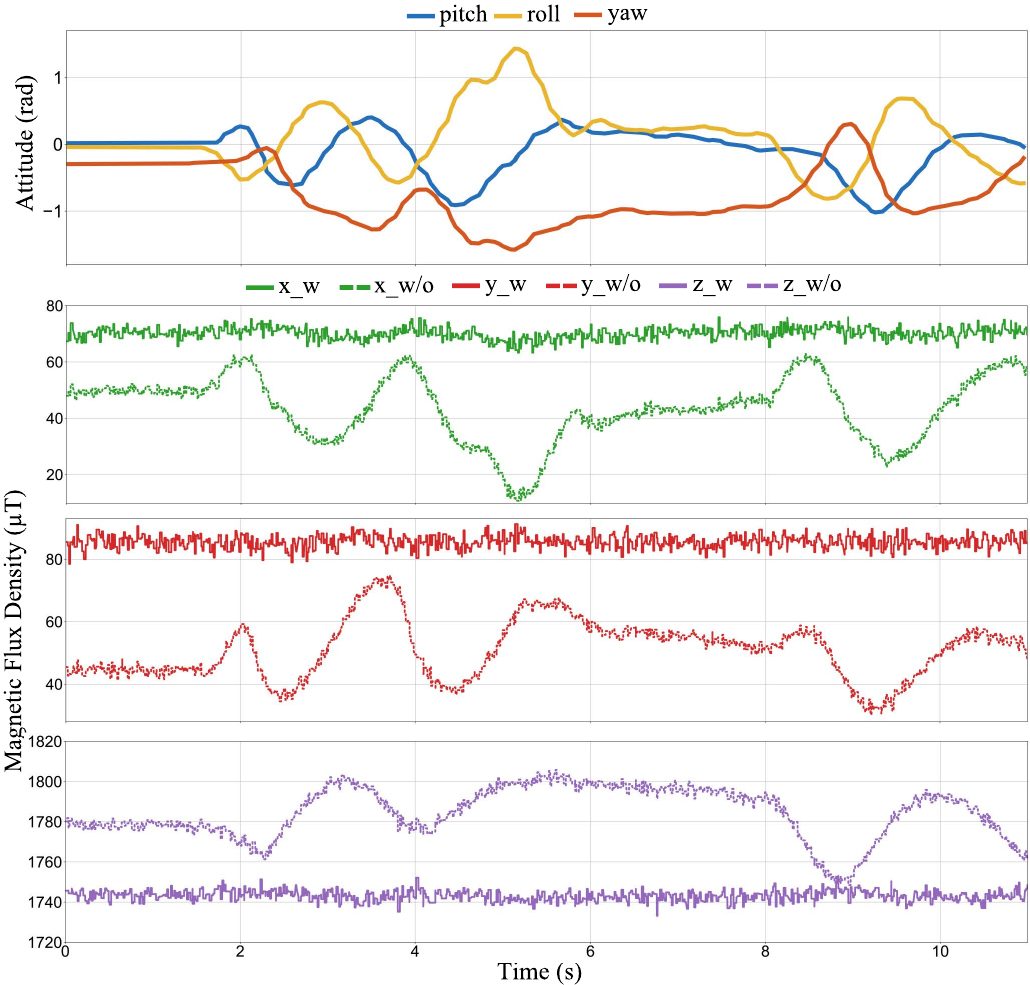}
\caption{Magnetic flux density measurements with/without geomagnetic compensation across different attitudes, showing reduced variation after compensation.}
\label{fig:mag_compensate}
\vspace{-0.3cm}
\end{figure}

Fig. \ref{fig:mag_compensate} compares magnetic flux density measurements with and without geomagnetic compensation. The uncompensated data exhibit significant attitude-dependent variations, while the compensated values maintain stability near baseline levels under no-load conditions.

\section{FORCE-AWARE GRASPING CONTROL}
\subsection{Position Control}
The overall control architecture, illustrated in Fig.~\ref{fig:pipeline}, integrates three main components: a position controller, an attitude controller, and a grasping force controller. The position controller employs an admittance control scheme to regulate the quadrotor's translational motion in response to external forces, balancing position tracking accuracy with compliance during physical interactions.

The admittance dynamics are described as:
\begin{align}
\bm{M}(\bm{\ddot{p}_{d}}-\bm{\ddot{p}_{r}}) + \bm{D}(\bm{\dot{p}_{d}}-\bm{\dot{p}_{r}}) + \bm{K}(\bm{p_{d}}-\bm{p_{r}}) = \hat{\bm{f}}_{ext},
\label{eq:admittance_position}
\end{align}
where $\bm{M}, \bm{D}, \bm{K}  \in \mathbb{R}^{3\times3}$ denote the virtual inertia, damping, and stiffness matrices respectively, $\bm{p_{r}}$, $\bm{\dot{p}_{r}}$ and $\bm{\ddot{p}_{r}}$ represent the reference trajectory components, and $\hat{\bm{f}}_{ext}$ is the estimated external force from tactile sensors.

The desired acceleration from the admittance controller is computed as:
\begin{equation}
\bm{\ddot{p}_{d}} = \bm{\ddot{p}_{r}} + \bm{M}^{-1}\Big(-\bm{D}(\bm{\dot{p}_{d}}-\bm{\dot{p}_{r}}) - \bm{K}(\bm{p_{d}}-\bm{p_{r}}) + \hat{\bm{f}}_{ext}\Big).
\end{equation}

A cascaded PD controller then computes the commanded acceleration:
\begin{equation}
\bm{a}_{cmd} = \bm{\ddot{p}_{d}} + \bm{K_{v}}(\bm{\dot{p}_{d}}-\bm{\dot{p}}) + \bm{K_{p}}(\bm{p_{d}}-\bm{p}),
\end{equation}
where $\bm{K_{v}}$ and $\bm{K_{p}}$ are diagonal gain matrices, $\bm{p}$ and $\bm{\dot{p}}$ are the measured position and velocity.

The external force is estimated by transforming sensor measurements from body frame $\mathcal{B}$ to world frame $\mathcal{W}$:
\begin{align}
\hat{\bm{f}}_{ext} = {^{\mathcal{W}}_{\mathcal{B}}\bm{R}}\sum_{i=1}^n ({^{\mathcal{B}}_{\mathcal{S}_{i}}\bm{R}\hat{\bm{f}_i}}),
\end{align}
where $\hat{\bm{f}_i}$ is the force measurement from the $i$-th sensor, ${^\mathcal{B}_{\mathcal{S}_{i}}\bm{R}}$ is the rotation matrix from the $i$-th sensor frame to the body frame, and ${^{\mathcal{W}}_{\mathcal{B}}\bm{R}}$ is the rotation matrix from the body frame to the world frame obtained from the state estimation.

Finally, the desired thrust is calculated as:
\begin{align}
\bm{T_{d}} = {^{\mathcal{W}}_{\mathcal{B}}\bm{R}^\top}(m\bm{a}_{cmd} - m\bm{g}),
\end{align}
where $m$ is the quadrotor mass and $\bm{g}$ is the gravity vector.

\subsection{Attitude Control}
A cascaded PID controller is implemented for attitude stabilization. The desired angular velocity $\bm{\omega_d}$ and angular velocity error $\bm{\omega_e}$ are defined as follows:
\begin{equation}
\label{eqn:omega_e}
\bm{\omega_d} = \bm{K_{R}} \cdot \bm{R_e},
\quad
\bm{\omega_e} = \bm{\omega_d} - \bm{\omega},
\end{equation}
where $\bm{K_{R}}$ is a positive definite gain matrix, $\bm{R_e}$ is the attitude error derived from reference and measured rotation matrices, and $\bm{\omega}$ represents the measured angular velocity from onboard IMU.

The desired torque $\bm{\tau_d}$ is computed using:
\begin{equation}
\label{eqn:tau_d}
\bm{\tau_d} = \bm{K_{P,\omega}} \bm{\omega_e} + \bm{K_{I,\omega}} \int \bm{\omega_e} dt + \bm{K_{D,\omega}} {\bm{\dot{\omega}_e}},
\end{equation}
where $\bm{K_{P,\omega}}$, $\bm{K_{I,\omega}}$, and $\bm{K_{D,\omega}}$ are positive definite gain matrices.

\begin{figure*}[htbp]
    \centering
    \includegraphics[width=0.92\linewidth]{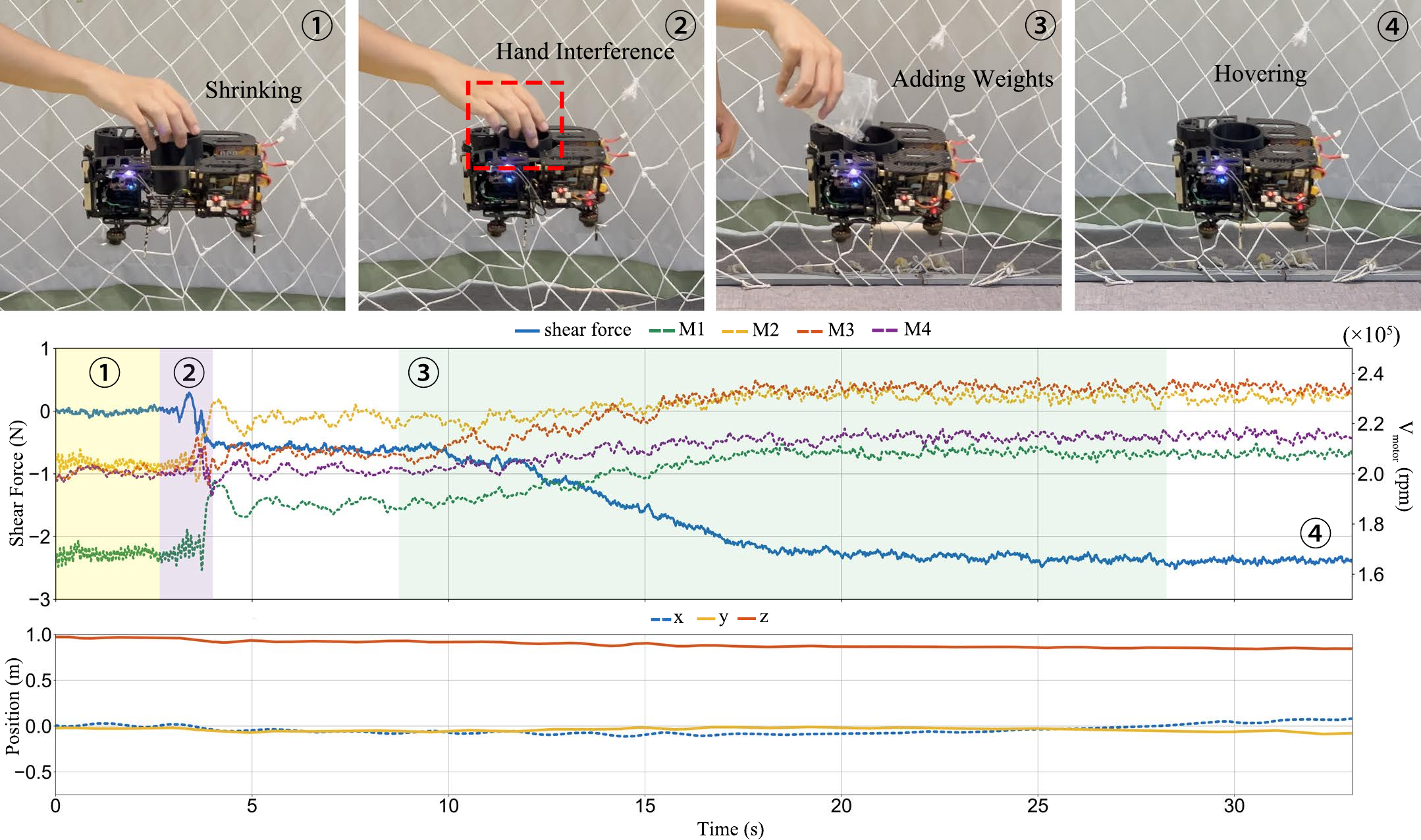}
    \caption{Experiment under dynamically changing load conditions. (1) The quadrotor shrinking to grasp a $50g$ 3D-printed PLA container. (2) The quadrotor tightened and grasped the container, showing force fluctuations during human hand release. (3) Adding glass beads to simulate real-time load variation. (4) The quadrotor maintaining stable hover under $230g$ additional load.}
    \label{fig:exp_dynamic}
    \vspace{-0.2cm}
\end{figure*}

\begin{figure*}[htbp]
    \centering
    \includegraphics[width=0.92\linewidth]{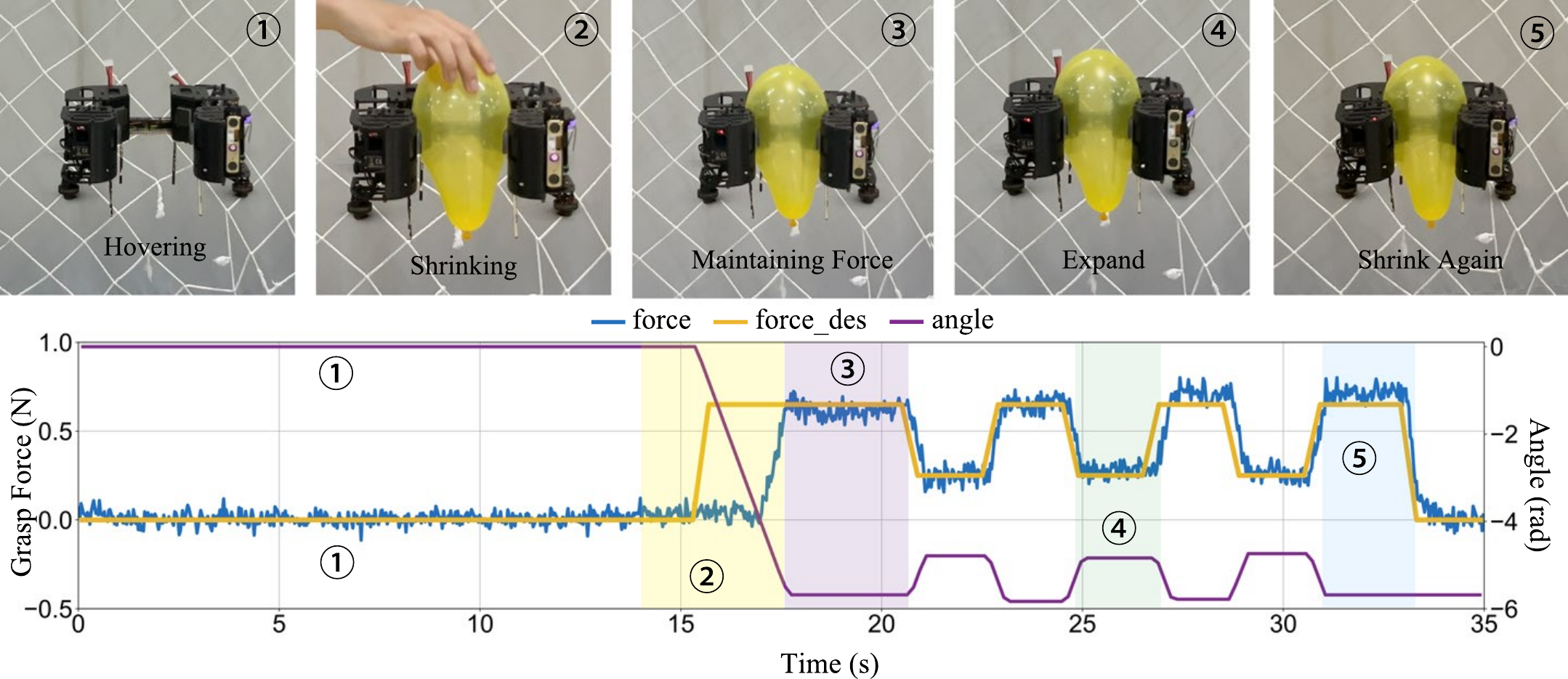}
    \caption{Closed-loop force control during balloon grasping. (1) The quadrotor maintains hover while the measured grasping force remains near zero. (2) The quadrotor begins to shrink and increases applied grasping force. (3) and (5) The quadrotor maintains a grasping force around $0.65N$, while (4) regulates near $0.25N$, showing larger balloon deformation under higher force compared to the smaller deformation at lower force.}
    \label{fig:exp_balloon}
    \vspace{-0.5cm}
\end{figure*}

\subsection{Grasp Control}
The grasping mechanism is implemented through servo motor angular velocity control. The total normal grasping force $\hat{f_g}$ is calculated by summing the normal force components from all individual sensor measurements:
\begin{align}
\hat{f_g} = \sum_{i=1}^n \norm{\hat{\bm{f}_i} \cdot \hat{\bm{n}_i}},
\end{align}
where $\hat{\bm{n}_i}$ represents the contact surface normal vector of the $i$-th sensor. To achieve the desired grasping force $f_d$, we employ an admittance control strategy described by the second-order dynamics:
\begin{align}
M_{\Delta \theta}\ddot{\Delta \theta}+B_{\Delta \theta}\dot{\Delta \theta}+K_{\Delta \theta} {\Delta\theta} = f_d - \hat{f_g},
\end{align}
where $M_{\Delta \theta}$, $B_{\Delta \theta}$, and $K_{\Delta \theta}$ represent the virtual inertia, damping, and stiffness parameters respectively, $\Delta \theta$ is the angular displacement from the reference position, and $f_d - \hat{f_g}$ is the force error between the desired and measured grasping forces. The target servo position $\theta_d$ is then computed by:
\begin{align}
\theta_d = \theta_{r} + \Delta \theta,
\end{align}
where $\theta_{r}$ denotes the initial servo angle position before contact is established. This admittance-based approach enables compliant grasping by generating appropriate angular adjustments $\Delta \theta$ in response to force errors, ensuring stable and adaptive interaction with objects of varying stiffness and geometry.

\begin{figure}[h]
	\centering
    \includegraphics[width= \linewidth]{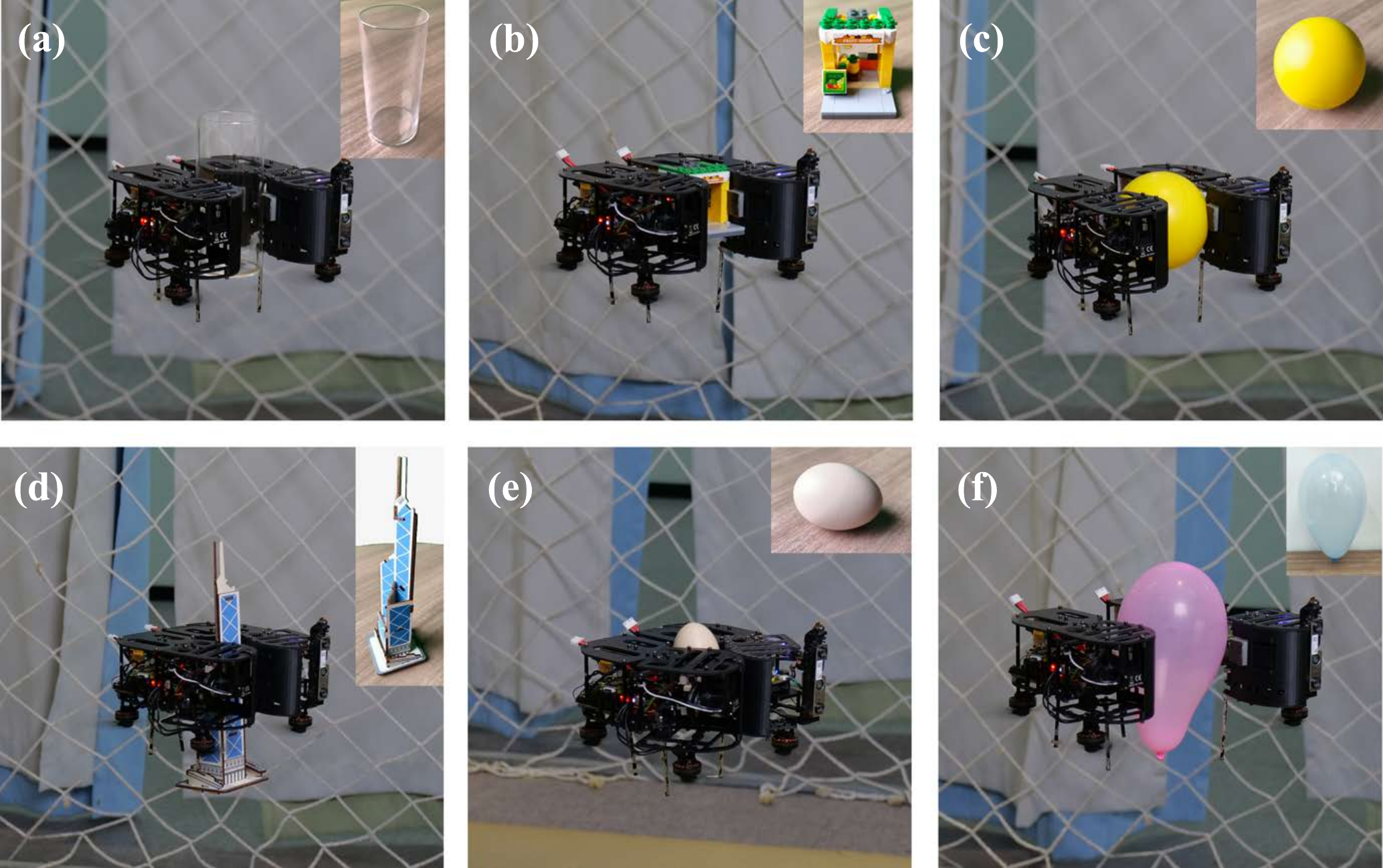}
    \caption{Grasping performance evaluation with various objects: (a) glass cup, (b) toy blocks, (c) stress ball, (d) wooden puzzle, (e) egg, and (f) balloon, demonstrating the system's adaptability to objects of different shapes, sizes, and fragility levels.}
	\label{fig:grasping}
    \vspace{-0.4cm}
\end{figure}

\section{EXPERIMENT}
This section presents experimental validation of the proposed force-aware control framework, focusing on three key capabilities: precise force regulation for fragile object manipulation, stability maintenance under dynamic payload variations, and robust aerial interaction without external infrastructure. All experiments were conducted on a custom deformable quadrotor equipped with six magnetic tactile sensors for multi-directional force sensing and visual-inertial odometry (VIO) for onboard localization, operating completely autonomously without motion-capture systems or offboard computation. 

Two main experiments are designed to evaluate different capabilities of the proposed controller: a balloon grasping task to validate precise force control performance with fragile objects, and a dynamic load variation test to assess stability maintenance under changing payload conditions, which collectively demonstrate the effectiveness of our force-aware control framework in various aerial manipulation scenarios.

\subsection{Sensor Setup}
The operational range of the sensor is defined within a spatial region of $x, y = \pm 3 \text{mm}$ and $z = -10 \pm 2 \text{mm}$. Within this working volume, the compensation coefficients for equation (\ref{eq:z_compensate}) were optimized and determined using \textit{Mathematica}, yielding the values $k_1 = 2.27851$, $c_1 = 0.0010535$, $k_2 = 0.878725$, and $c_2 = -0.0000785667$. Prior to flight experiments, parameter $\bm{a}$ is calibrated using a commercial force-torque sensor. During system initialization, parameter $\bm{b}$ is obtained by averaging multiple sensor readings under no-load conditions and applying a zero-offset correction procedure. The three-dimensional force vector can then be robustly estimated via the decoupling model given in equation (\ref{eq:force_compensate}).


\subsection{Dynamic Load Condition}
To simulate real-time load variation, we designed a lightweight $50g$ PLA 3D-printed container and used $5mm$ diameter glass beads as an adjustable payload. During the experiment, the quadrotor first grasps the empty container while a human operator gradually adds approximately $180g$ of glass beads to increase the load dynamically. As shown in Fig. \ref{fig:exp_dynamic}, the system initially detects a shear force of approximately $0.60N$ and responds by increasing rotor speed. As additional beads are poured into the container, the quadrotor continuously adjusts thrust to maintain stability, with the sensor finally recording a stable shear force of $2.37N$. Rotor speed adjustments show strong correlation with load variations, demonstrating the force-aware control system's capability to respond rapidly to dynamic payload changes. The experimental results in Fig. \ref{fig:exp_dynamic} also indicate that the quadrotor maintains relatively stable position control with minimal deviation, while the altitude remains essentially level despite the increasing load. This experiment validates the system's ability to maintain stable flight under dynamically changing load conditions through real-time force-aware control.

\subsection{Diverse Object Grasping}
As illustrated in Fig. \ref{fig:grasping}, we evaluated the grasping performance on several common fragile objects, including balloons, eggs, toy blocks, stress ball, wooden puzzle and glass cup. These objects cover a wide range of material and geometric properties—from soft, deformable to rigid yet fragile—representing common real-world grasping challenges. For instance, balloons require gentle handling to avoid bursting from excessive compression; eggs demand high force-control precision to prevent cracks; and glass cup necessitate a balance between grip force and stability to prevent breakage. The selection of these diverse objects effectively validates the general applicability of our system. Benefiting from the soft and compliant structure of our tactile sensor, our approach significantly reduces impact forces during grasping and enables safe physical interaction with fragile objects. 

To further validate the precise force control capability of our system, we conducted a balloon grasping experiment that demonstrates fragile manipulation under critical force constraints. As shown in Fig. \ref{fig:exp_balloon}, the quadrotor continuously regulates the servo position through closed-loop force feedback to track the desired contact force. The adaptive adjustment maintains the applied force within a safe range of $0.25$-$0.65N$, resulting in controlled elastic deformation of the balloon without causing rupture or excessive compression. This scenario particularly highlights the system's capability in handling deformable objects requiring high-level force sensitivity, validating the effectiveness of our force-aware control framework for robust aerial manipulation tasks.

\begin{figure}[t]
	\centering
    \includegraphics[width=\linewidth]{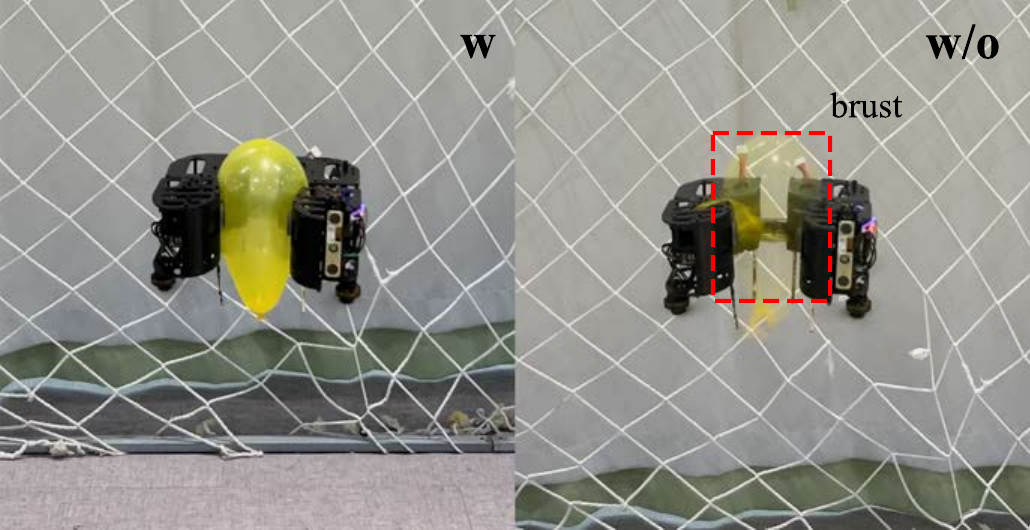}
    \caption{Ablation study of grasping a balloon. Left: Our controller successfully grasps the balloon without causing damage. Right: Under open-loop control (w/o feedback), excessive force causes the balloon to burst.}
	\label{fig:ablation_ballon}
    \vspace{-0.4cm}
\end{figure}

\begin{figure}[t]
	\centering
    \includegraphics[width= \linewidth]{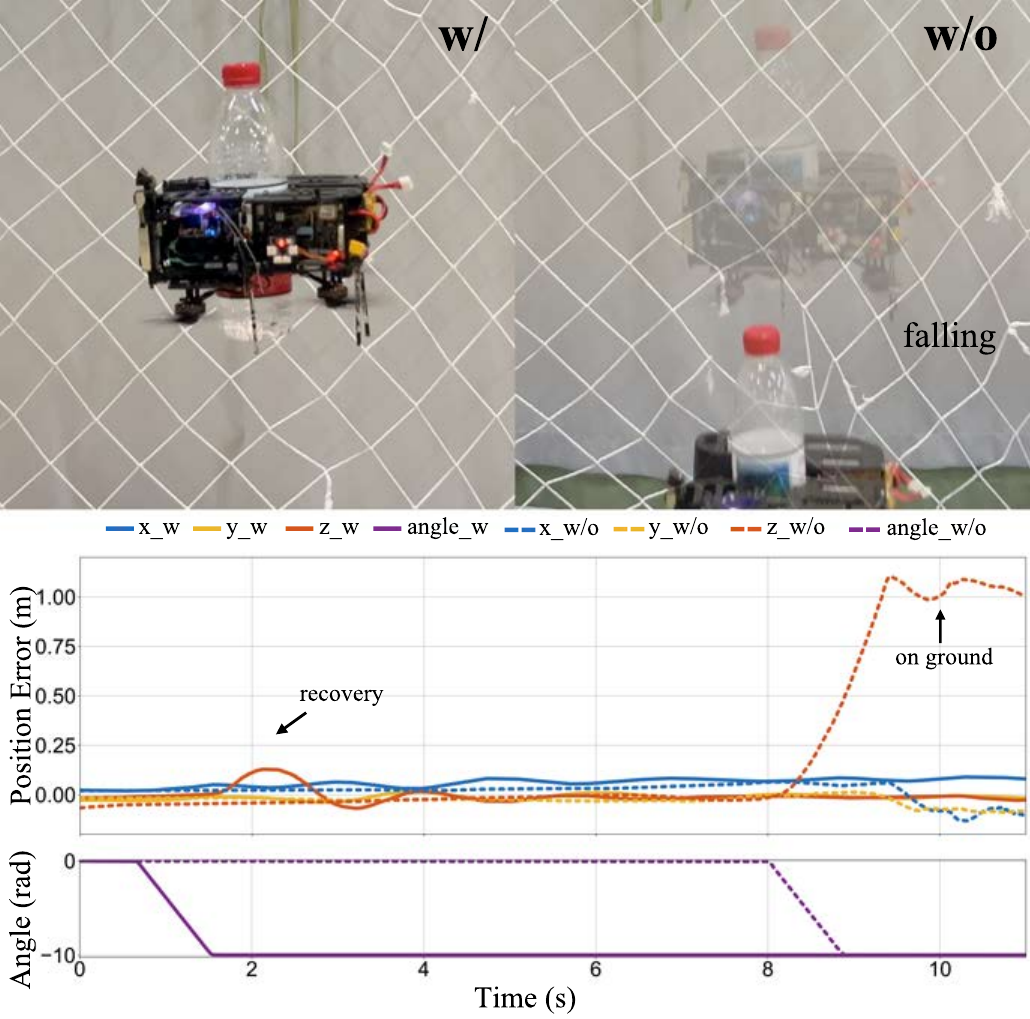}
    \caption{Ablation study on the proposed controller for grasping a $180g$ water bottle. Left: Our controller maintains a stable altitude after grasp. Right: The baseline (w/o proposed module) fails to maintain stability, resulting in the quadrotor falling to the ground.}
	\label{fig:ablation_bottle}
    \vspace{-0.5cm}
\end{figure}

\subsection{Ablation Studies}
We conducted ablation studies to validate the performance of the proposed framework. Two distinct grasping tasks were designed to evaluate different aspects of the system: one involved grasping a $180g$ water bottle to test weight-bearing capacity and stability under substantial load, while the other involved manipulating a deformable balloon to assess fine force control and compliance with fragile objects. Comparative results with (w/) and without (w/o) the proposed tactile feedback are presented in Fig. \ref{fig:ablation_ballon} and \ref{fig:ablation_bottle}. In the balloon grasping scenario, as shown in Fig. \ref{fig:ablation_ballon}, the quadrotor equipped with tactile sensing avoids excessive squeezing and prevents the balloon from bursting by regulating the gripper’s pressure in real time. In the water bottle grasping task, the absence of force feedback leads to insufficient thrust adjustment, causing the quadrotor to lose altitude continuously upon contact and eventually drop the object. Quantitative position error data clearly show that with feedback control, the height deviation is quickly compensated and stabilizes, whereas without feedback, the vertical error increases drastically until the quadrotor falls to ground. These experiments collectively demonstrate the necessity of the proposed sensing and control framework for achieving robust aerial manipulation, especially under varying load conditions and handling fragile objects.

\section{CONCLUSIONS}
In this work, we present a deformable quadrotor platform with integrated tactile sensing and force-based control, enabling in-flight force perception and real-time manipulation. The proposed framework is experimentally validated through the successful grasping of both weight-significant and fragile objects, demonstrating its practical applicability. These results underscore the potential of our method to support autonomous aerial manipulation in real-world scenarios. Future work will focus on developing grasp planning mechanisms to increase autonomy and investigating material recognition techniques to enhance the robustness of grasping operations.

\addtolength{\textheight}{-12cm}   


\footnotesize{
\bibliographystyle{IEEEtran}
\bibliography{main}

@article{lin20239dtact,
  title={9dtact: A compact vision-based tactile sensor for accurate 3d shape reconstruction and generalizable 6d force estimation},
  author={Lin, Changyi and Zhang, Han and Xu, Jikai and Wu, Lei and Xu, Huazhe},
  journal={IEEE Robotics and Automation Letters},
  volume={9},
  number={2},
  pages={923--930},
  year={2023},
  publisher={IEEE}
}

@article{lin2025pp,
  title={PP-Tac: Paper Picking Using Tactile Feedback in Dexterous Robotic Hands},
  author={Lin, Pei and Huang, Yuzhe and Li, Wanlin and Ma, Jianpeng and Xiao, Chenxi and Jiao, Ziyuan},
  journal={arXiv preprint arXiv:2504.16649},
  year={2025}
}

@article{liang2023adaptive,
  title={Adaptive force tracking impedance control for aerial interaction in uncertain contact environment using barrier function},
  author={Liang, Jiacheng and Zhong, Hang and Wang, Yaonan and Chen, Yanjie and Zeng, Junhao and Mao, Jianxu},
  journal={IEEE Transactions on Automation Science and Engineering},
  year={2023},
  publisher={IEEE}
}

@article{yuan2017gelsight,
  title={Gelsight: High-resolution robot tactile sensors for estimating geometry and force},
  author={Yuan, Wenzhen and Dong, Siyuan and Adelson, Edward H},
  journal={Sensors},
  volume={17},
  number={12},
  pages={2762},
  year={2017},
  publisher={MDPI}
}

@article{xue2025reactive,
  title={Reactive diffusion policy: Slow-fast visual-tactile policy learning for contact-rich manipulation},
  author={Xue, Han and Ren, Jieji and Chen, Wendi and Zhang, Gu and Fang, Yuan and Gu, Guoying and Xu, Huazhe and Lu, Cewu},
  journal={arXiv preprint arXiv:2503.02881},
  year={2025}
}

@article{guo2024flying,
  title={Flying calligrapher: Contact-aware motion and force planning and control for aerial manipulation},
  author={Guo, Xiaofeng and He, Guanqi and Xu, Jiahe and Mousaei, Mohammadreza and Geng, Junyi and Scherer, Sebastian and Shi, Guanya},
  journal={IEEE Robotics and Automation Letters},
  year={2024},
  publisher={IEEE}
}

@article{wu2026hand,
  title={Hand-like autonomous flying robot for airborne grasping and interaction},
  author={Wu, Yuze and Yang, Fan and Jin, Rui and Zhong, Yuhang and Wang, Junjie and Wu, Xuankang and Gao, Fei},
  journal={Nature Communications},
  year={2026},
  publisher={Nature Publishing Group UK London}
}

@article{bodie2019omnidirectional,
  title={An omnidirectional aerial manipulation platform for contact-based inspection},
  author={Bodie, Karen and Brunner, Maximilian and Pantic, Michael and Walser, Stefan and Pf{\"a}ndler, Patrick and Angst, Ueli and Siegwart, Roland and Nieto, Juan},
  journal={arXiv preprint arXiv:1905.03502},
  year={2019}
}

@article{benzi2022adaptive,
  title={Adaptive tank-based control for aerial physical interaction with uncertain dynamic environments using energy-task estimation},
  author={Benzi, Federico and Brunner, Maximilian and Tognon, Marco and Secchi, Cristian and Siegwart, Roland},
  journal={IEEE Robotics and Automation Letters},
  volume={7},
  number={4},
  pages={9129--9136},
  year={2022},
  publisher={IEEE}
}

@article{huang20243d,
  title={3d-vitac: Learning fine-grained manipulation with visuo-tactile sensing},
  author={Huang, Binghao and Wang, Yixuan and Yang, Xinyi and Luo, Yiyue and Li, Yunzhu},
  journal={arXiv preprint arXiv:2410.24091},
  year={2024}
}

@article{bhirangi2021reskin,
  title={Reskin: versatile, replaceable, lasting tactile skins},
  author={Bhirangi, Raunaq and Hellebrekers, Tess and Majidi, Carmel and Gupta, Abhinav},
  journal={arXiv preprint arXiv:2111.00071},
  year={2021}
}

@article{lin2025float,
  title={FLOAT Drone: A Fully-actuated Coaxial Aerial Robot for Close-Proximity Operations},
  author={Lin, Junxiao and Ji, Shuhang and Wu, Yuze and Wu, Tianyue and Han, Zhichao and Gao, Fei},
  journal={arXiv preprint arXiv:2503.00785},
  year={2025}
}

@article{wu2023ring,
  title={Ring-rotor: A novel retractable ring-shaped quadrotor with aerial grasping and transportation capability},
  author={Wu, Yuze and Yang, Fan and Wang, Ze and Wang, Kaiwei and Cao, Yanjun and Xu, Chao and Gao, Fei},
  journal={IEEE Robotics and Automation Letters},
  volume={8},
  number={4},
  pages={2126--2133},
  year={2023},
  publisher={IEEE}
}

@article{ubellacker2024high,
  title={High-speed aerial grasping using a soft drone with onboard perception},
  author={Ubellacker, Samuel and Ray, Aaron and Bern, James M and Strader, Jared and Carlone, Luca},
  journal={npj Robotics},
  volume={2},
  number={1},
  pages={5},
  year={2024},
  publisher={Nature Publishing Group UK London}
}

@article{xu2023aerial,
  title={Aerial shooting manipulator for distant grasping},
  author={Xu, Mengxin and Huang, Siyuan and He, Ruokun and Yu, Dafang and Wang, Hesheng},
  journal={IEEE Robotics and Automation Letters},
  volume={8},
  number={4},
  pages={1991--1998},
  year={2023},
  publisher={IEEE}
}

@article{bhirangi2024anyskin,
  title={Anyskin: Plug-and-play skin sensing for robotic touch},
  author={Bhirangi, Raunaq and Pattabiraman, Venkatesh and Erciyes, Enes and Cao, Yifeng and Hellebrekers, Tess and Pinto, Lerrel},
  journal={arXiv preprint arXiv:2409.08276},
  year={2024}
}

@article{zhao2023versatile,
  title={Versatile articulated aerial robot DRAGON: Aerial manipulation and grasping by vectorable thrust control},
  author={Zhao, Moju and Okada, Kei and Inaba, Masayuki},
  journal={The International Journal of Robotics Research},
  volume={42},
  number={4-5},
  pages={214--248},
  year={2023},
  publisher={SAGE Publications Sage UK: London, England}
}

@article{li2025aerothrow,
  title={AeroThrow: An Autonomous Aerial Throwing System for Precise Payload Delivery},
  author={Li, Ziliang and Chen, Hongming and Lin, Yiyang and Ye, Biyu and Lyu, Ximin},
  journal={arXiv preprint arXiv:2507.13903},
  year={2025}
}

@article{dai2024split,
  title={Split-type magnetic soft tactile sensor with 3D force decoupling},
  author={Dai, Huangzhe and Zhang, Chengqian and Pan, Chengfeng and Hu, Hao and Ji, Kaipeng and Sun, Haonan and Lyu, Chenxin and Tang, Daofan and Li, Tiefeng and Fu, Jianzhong and others},
  journal={Advanced Materials},
  volume={36},
  number={11},
  pages={2310145},
  year={2024},
  publisher={Wiley Online Library}
}

@article{wang2025safe,
  title={Safe and Agile Transportation of Cable-Suspended Payload via Multiple Aerial Robots},
  author={Wang, Yongchao and Wang, Junjie and Zhou, Xiaobin and Yang, Tiankai and Xu, Chao and Gao, Fei},
  journal={arXiv preprint arXiv:2501.15272},
  year={2025}
}

@inproceedings{appius2022raptor,
  title={Raptor: Rapid aerial pickup and transport of objects by robots},
  author={Appius, Aurel X and Bauer, Erik and Bl{\"o}chlinger, Marc and Kalra, Aashi and Oberson, Robin and Raayatsanati, Arman and Strauch, Pascal and Suresh, Sarath and von Salis, Marco and Katzschmann, Robert K},
  booktitle={2022 IEEE/RSJ International Conference on Intelligent Robots and Systems (IROS)},
  pages={349--355},
  year={2022},
  organization={IEEE}
}

@article{tzoumanikas2020aerial,
  title={Aerial manipulation using hybrid force and position nmpc applied to aerial writing},
  author={Tzoumanikas, Dimos and Graule, Felix and Yan, Qingyue and Shah, Dhruv and Popovic, Marija and Leutenegger, Stefan},
  journal={arXiv preprint arXiv:2006.02116},
  year={2020}
}
}

\end{document}